\documentclass{ecai}
\usepackage{times}
\usepackage{graphicx}
\usepackage{latexsym}
\usepackage{subfigure}
\usepackage{booktabs} 
\usepackage{hyperref}
\newcommand{\ra}[1]{\renewcommand{\arraystretch}{#1}}

\ecaisubmission   

\usepackage{amsmath}
\usepackage{amsfonts}
\DeclareMathOperator*{\KL}{KL}%
\DeclareMathOperator*{\diag}{diag}%
\newcommand{\nn}{\nonumber \\}%
\newcommand{\mE}{\mathbb{E}}
\newcommand{\mH}{\mathbb{H}}
\newcommand{\mI}{\mathbb{I}}
\newcommand{\mL}{\mathcal{L}}
\newcommand{\mN}{\mathcal{N}}
\newcommand{\mR}{\mathbb{R}}
\newcommand{\mS}{\mathcal{S}}
\newcommand{\mU}{\mathcal{U}}

\newcommand{\bc}{\mathbf{c}}
\newcommand{\bu}{\mathbf{u}}
\newcommand{\bx}{\mathbf{x}}
\newcommand{\bz}{\mathbf{z}}

\usepackage{xcolor}
\newif\ifShowComments
\ShowCommentstrue 
\begin{document}

\title{Improving VAE generations of multimodal data through data-dependent conditional priors}

\author{Frantzeska Lavda\textsuperscript{*}\institute{University of Geneva,
Switzerland, email: frantzeska.lavda@etu.unige.ch} \institute{Geneva School of Business Administration (DMML Group), HES-SO, Switzerland, email: name.surname@hesge.ch, \newline * equal contribution} \and Magda Gregorov\'a\textsuperscript{*}\textsuperscript{2} \and Alexandros Kalousis\textsuperscript{2} }

\maketitle
\bibliographystyle{ecai}

\begin{abstract}
One of the major shortcomings of variational autoencoders is the inability to produce generations from the individual modalities of data originating from mixture distributions.
This is primarily due to the use of a simple isotropic Gaussian as the prior for the latent code in the ancestral sampling procedure 
for the data generations.
We propose a novel formulation of variational autoencoders, conditional prior VAE (CP-VAE), which learns to differentiate between the individual mixture components and therefore allows for generations from the distributional data clusters.
We assume a two-level generative process with a continuous (Gaussian) latent variable sampled conditionally on a discrete (categorical) latent component.
The new variational objective naturally couples the learning of the posterior and prior conditionals, and the learning of the latent categories encoding the multimodality of the original data in an unsupervised manner.
The data-dependent conditional priors are then used to sample the continuous latent code when generating new samples from the individual mixture components corresponding to the multimodal structure of the original data. 
Our experimental results illustrate the generative performance of our new model comparing to multiple baselines.
\end{abstract}

\section{Introduction}\label{sec:Intro}

Variational autoencoders (VAEs) \cite{kingmaAutoEncodingVariationalBayes2014,rezendeStochasticBackpropagationApproximate2014} are deep generative models for learning complex data distributions.
They consist of an encoding and a decoding network parametrizing the variational approximate posterior and the conditional data distributions in a latent variable generative model.

Though powerful and theoretically elegant, the VAEs in their basic form suffer from multiple deficiencies stemming from the mathematically convenient yet simplistic distributional assumptions.
Multiple strategies have been proposed to increase the richness or the interpretability of the latent code \cite{rezendeVariationalInferenceNormalizing2015,burdaImportanceWeightedAutoencoders2016,kingmaImprovedVariationalInference2016,chenVariationalLossyAutoencoders2017,jiangVariationalDeepEmbedding2017,nalisnickStickBreakingVariationalAutoencoders2017,zhaoInfoVAEInformationMaximizing2017,alemiFixingBrokenELBO2018,davidsonHypersphericalVariationalAutoEncoders2018,daiDiagnosingEnhancingVAE2019}.
These mostly argue for more flexible posterior inference procedure or for the use of more complex approximate posterior distributions to facilitate the encoding of non-trivial data structures within the latent space.

In this paper we reason that for generating realistic samples of data originating from complex distributions, it is the prior which lacks expressiveness.
Accordingly, we propose a new VAE formulation, conditional prior VAE (CP-VAE), with two-level hierarchical generative model combining a categorical and a continuous (Gaussian) latent variables.

The hierarchical conditioning of the continuous latent variable on the discrete latent component
is particularly suitable for modelling multimodal data distributions such as distributional mixtures.
Importantly, it also gives us better control of the procedure for generating new samples.
Unlike in the standard VAE, we can sample data from specific mixture components at will.
This is particularly critical if the generative power of VAEs shall be used in conjunction with methods requiring the identification of the distributional components such as in continual learning \cite{ramapuramLifelongGenerativeModeling2019,lavdaContinualClassificationLearning2018a}.

As recently shown \cite{daiDiagnosingEnhancingVAE2019,locatelloChallengingCommonAssumptions2019a}, without supervision (as in our setting), enforcing independence factorization in the latent space does not guarantee recovering meaningful sources of variation in the original space.
Therefore in our CP-VAE formulation, we let the model to fully utilize the capacity of the latent space by allowing for natural conditional decomposition in the generative and inference graphical models.

We formulate the corresponding variational lower bound on the data log-likelihood and use it as the optimization objective in the training.
In the spirit of empirical Bayes, we propose to estimate the parameters of the conditional priors from the data together with the parameters of the variational posteriors in a joint learning procedure.
This ensures that the inferred structure of the latent space can be exploited in the data generations.

\section{Variational autoencoders}\label{sec:VAE}

Variational autoencoders (VAEs) \cite{kingmaAutoEncodingVariationalBayes2014,rezendeStochasticBackpropagationApproximate2014} are deep Bayesian generative models relying on the principals of amortized variational inference to approximate the complex distributions $p(\bx)$ from which the observed data $\mS = \{\bx_i\}_{i=1}^n$ originate.

In their basic form, they model the unknown ground-truth $p(\bx)$ by a parametric distribution $p_\theta(\bx)$ with a latent variable generative process
\begin{equation}\label{eq:vaeGenerativeModel}
p_\theta(\bx) = \int p_\theta(\bx | \bz) p(\bz) dz \enspace .
\end{equation}
The typical assumptions are of a simple isotropic Gaussian prior $p(\bz)$ for the latent variable $\bz$ and, depending on the nature of the data $\bx$, factorized Bernoulli or Gussian distributions for the data conditionals $p_\theta(\bx | \bz)$.
These per-sample conditionals are parametrized by a deep neural network, a \emph{decoder}.
Once the decoder network is properly trained, we can sample new data examples from the learned data distribution $p_\theta(\bx)$ by ancestral sampling procedure: sample the latent $\bz$ from the prior $p(\bz)$ and pass it through the stochastic \emph{decoder} $p_\theta(\bx | \bz)$ to obtain the sample $\bx$. 

The VAEs employ the strategy of amortized variational inference.
They approximate the intractable posteriors $p_\theta(\bz | \bx)$ by factorized Gaussian distributions $q_\phi(\bz | \bx)$ and infer the variational parameters $\phi$ of the approximate per-sample posteriors through a deep neural network, an \emph{encoder}.

The encoder and decoder networks are trained end-to-end by stochastic gradient-based optimization maximizing the sample estimate of lower bound $\mL_{\theta, \phi} = \mE_{p(\bx)} \mL_{\theta, \phi}(\bx)$ on the data log-likelihood
\begin{gather}\label{eq:vaeElbo}
\frac{1}{n} \sum_i^n \log p_\theta(\bx_i) \geq \frac{1}{n} \sum_i^n \mL_{\theta, \phi}(\bx_i) \approx \mL_{\theta, \phi} \nn
\mL_{\theta, \phi}(\bx) = \underbrace{\mE_{q_\phi(\bz | \bx)} \log p_\theta(\bx | \bz)}_A
- \underbrace{\KL\left(q_\phi(\bz | \bx) || p(\bz)\right)}_B \enspace ,
\end{gather}
where $\KL(.||.)$ is the Kullback-Leibler divergence.
The first term $A$ in equation \eqref{eq:vaeElbo} can be seen as a negative reconstruction cost, term $B$ penalizes the deviations of the approximate posterior from the fixed prior $p(\bz)$ and has a regularizing effect on the model learning.

The gradients of the lower bound with respect to the model parameters $\theta$ can be obtained streighforwardly through Monte Carlo estimation.
For the posterior parameters $\phi$, the gradients are estimated by stochastic backpropagation using a location-scale transformation known as the reparametrization trick.

\section{VAE with data-dependent conditional priors}\label{sec:CondVAE}

The mathematically and practically convenient assumption of the factorial Gaussian approximate posterior $q_\phi(\bz | \bx)$ has been previously contested as one of the major limitations of the basic VAE architecture.
For complex data distributions $p(\bx)$ the simple Gaussian $q_\phi(\bz | \bx)$ may not be flexible enough to approximate well the true posterior $p_\theta(\bz | \bx)$.
Scalable methods for enriching the posterior distributions through variable transformations have been proposed \cite{rezendeVariationalInferenceNormalizing2015,kingmaImprovedVariationalInference2016,bergSylvesterNormalizingFlows2019} alongside alternatives for improving the inference procedure
\cite{burdaImportanceWeightedAutoencoders2016,zhaoInfoVAEInformationMaximizing2017}.

These methods improve rather dramatically the lower bound and log-likelihood maximization objectives by learning latent representations more appropriate for the complex data structures.
However, these improvements cannot be translated into better generations without a closer match between the posterior and prior distributions used for sampling the latent variables during inference and data generations respectively.

We propose a new VAE formulation, conditional prior VAE (CP-VAE), with a conditionally structured latent representation which encourages a better match between prior and posterior distributions by jointly learning their parameters from the data.

\subsection{Two-level generative process}\label{sec:condVAE_genProcess}

We consider a two-level hierarchical generative process for the observed data
\begin{equation}\label{eq:condVaeGenerativeModel}
p_\theta(\bx) = \sum_c \int p_\theta(\bx | \bz, \bc) p_\varphi(\bz | \bc) p(\bc) dz \enspace .
\end{equation}
The latent space is composed of a continuous $\bz$ and a discrete $\bc$ component with a joint prior distribution
\begin{equation}\label{eq:condVaePrior}
p(\bz , \bc) = p_\varphi(\bz | \bc) p(\bc) \enspace .
\end{equation}
We assume a uniform categorical as a prior distribution for the discrete component $\bc$ so that for each of the $K$ categories $p(c_k) = 1/K, \ k = 1, \ldots, K$.
The conditionals of the continuous component are factorised Gaussians\footnote{Full-covariance Gaussian could be parametrized through a lower-triangular matrix corresponding to the Cholesky decomposition at an increased computational cost \cite[section 2.6.1]{kingmaVariationalInferenceDeep2017}.}
\begin{equation}\label{eq:condVaeContiPrior}
p_\varphi(\bz | c_k) = \prod_i p_\varphi(z_i | c_k) = \prod_i \mN\left(z_i \,|\, \mu_{ik}, \sigma_{ik}^2)\right), \ k = 1, \ldots, K \enspace . 
\end{equation}

The compositional prior we propose is well suited for generations of new samples from multimodal data distributions mixing multiple distributional components. 
In contrast to sampling from a simple isotropic Gaussian prior which concentrates symmetrically around the origin, we can sample the latent code from discontinuous parts of the latent space.
These are expected to represent data clusters corresponding to the originating distributional mixing.

In addition, the variations encoded into the continous part of the latent space are also sampled conditionally and therefore are better adapted to represent the important factors of data variations within the distributional clusters.
This is in contrast to the single common continuous distribution of the basic VAE (section \ref{sec:VAE}) or VAEs with similar continuous-discrete composition of the latents as ours which, however, assume independence between the two parts of the latent representation \cite{dupontLearningDisentangledJoint2018}.

The data conditional $p_\theta(\bx | \bz, \bc)$ is parametrised by a \emph{decoder} network $d_\theta(\bz, \bc)$ as a Bernoulli$\left(\bx \,|\,  d_\theta(\bz, \bc) \right)$ or a Gaussian $\mN\left(\bx \,|\, d_\theta(\bz, \bc), \sigma^2 I \right)$ distribution depending on the nature of the data $\bx$.
\begin{figure}
\centerline{\includegraphics[width=0.9\columnwidth]{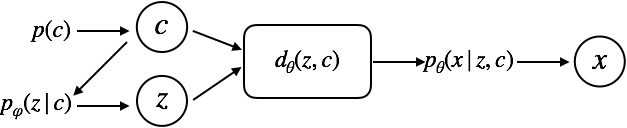}}
\caption{To generate new examples from the learned data distribution $p_\theta(\bx)$, we sample the discrete and continuous latent variables from the two-level prior and pass those through the decoder.} \label{fig:decoding}
\end{figure}

\subsubsection{Data-dependent conditional priors}

There is no straightforward way to fix the parameters $\varphi = (\mu, \sigma)$ in the distributions \eqref{eq:condVaeContiPrior} for each of the conditioning categories $c_k$ a priori.
Instead of placing hyper-priors on the parameters and expanding to full hierarchical Bayesian modelling, we estimate the prior parameters from the data through a relatively simple procedure resembling the empirical Bayes technique \cite{murphyMachineLearningProbabilistic2012}.

As explained in section \ref{sec:condVaeObjective}, the conditional $p_\varphi(\bz | \bc)$ enters our objective function through a $\KL$ divergence term.
The prior parameters $\varphi$ therefore can be optimized by backpropagation together with learning the encoder and decoder parameters $\phi$ and $\theta$. 
Once the model is trained, all the parameters are fixed and the learned prior $p_\varphi(\bz | \bc)$ can be used in the ancestral sampling procedure described above to generate new data samples.

\subsection{Inference model}

As in standard VAEs, we employ amortized variational inference to learn the unknown data distribution.
We use the approximate posterior distribution
\begin{equation}\label{eq:condVaePosterior}
q_\phi(\bz, \bc | \bx) = q_\phi(\bz| \bx, \bc) q_\phi(\bc| \bx) 
\end{equation}
in place of the intractable posterior $p_\theta(\bz, \bc | \bx)$.
Our approximate posterior replicates the two-level hierarchical structure of the prior.
In this way we ensure that the latent samples are structurally equivalent both during inference and new samples generations.
This is not the case in other hierarchical latent models which rely on simplifying mean field assumptions for the posterior inference \cite{jiangVariationalDeepEmbedding2017,dilokthanakulDeepUnsupervisedClustering2017}.

We use \emph{encoder} network with a gated layer $e_\phi(\bx) = (\pi_\phi(\bx), \mu_\phi(\bx, \pi), \sigma_\phi(\bx, \pi))$ for the amortized inference of the variational approximate posteriors 
\begin{gather*}
q_\phi(\bc| \bx) = \text{Cat}\left(\pi_\phi(\bx)\right) \\
q_\phi(\bz| \bx, \bc) = \mN(\mu_\phi(\bx, \pi), \diag(\sigma^2_\phi(\bx, \pi))) \enspace .
\end{gather*}
\begin{figure}
\centerline{\includegraphics[width=0.62\columnwidth]{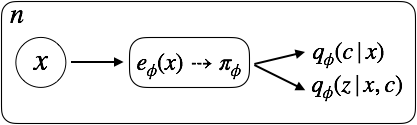}}
\caption{The encoder infers the parameters of the discrete and continuous approximate posteriors using a gated layer for the hierarchical conditioning.} \label{fig:decoding}
\end{figure}

\subsection{Optimization objective}\label{sec:condVaeObjective}

As customary in variational inference methods, our optimization objective is the maximization of the lower bound on the data log-likelihood
\begin{gather}\label{eq:condVaeElbo}
\mL_{\theta, \phi}(\bx) = \underbrace{\mE_{q_\phi(\bz, \bc | \bx)} \log p_\theta(\bx | \bz, \bc)}_A
- \underbrace{\KL\left(q_\phi(\bz, \bc | \bx) || p_\varphi(\bz, \bc)\right)}_B \enspace .
\end{gather}
This is a straightforward adaptation of the bound from equation \eqref{eq:vaeElbo} to the compositional latent code $(\bz, \bc)$ with similar interpretations for the $A$ and $B$ terms.
Using the posterior and prior distribution decompositions from equations \eqref{eq:condVaePrior} and \eqref{eq:condVaePosterior}, the $\KL$ term in $B$ can be rewritten as a sum of two $\KL$ divergences which are more amenable to practical implementation: $B1$ for the continuous conditional distributions and $B2$ for the discrete.
\begin{multline}\label{eq:condVaeKLs}
\underbrace{\KL\left(q_\phi(\bz, \bc | \bx) || p_\varphi(\bz, \bc)\right)}_B = \\
=
\underbrace{\KL\left(q_\phi(\bc| \bx) || p(\bc)\right)}_{B1}
+ \underbrace{\mE_{q_\phi(\bc|\bx)} \KL\left( q_\phi(\bz| \bx, \bc) || p_\varphi(\bz | \bc)\right)}_{B2}
\end{multline}

The minimization of the $\KL$ divergence between the categorical posterior and the fixed uniform prior in the first term $B1$ is equivalent to maximizing the entropy of the categorical posterior $\mH\left(q_\phi(\bc|\bx)\right)$ (up to a constant).
\begin{equation}\label{eq:condVae_KLB1}
\underbrace{\KL\left(q_\phi(\bc| \bx) || p(\bc)\right)}_{B1}
= - \mH\left(q_\phi(\bc|\bx)\right) + \log K
\end{equation}
The second term $B2$ can be seen as a weighted average of the $KL$ divergences between the posterior and prior conditionals.
The weights are the probabilities of the posterior categorical distribution so that the two conditionals are pushed together more strongly for those observations $\bx$ and latent categories $c_k$ to which the model assigns high probability.
The $KL$s can be conveniently evaluated in a closed form as both the posterior and the prior conditionals are diagonal Guassians.

We train the model by a stochastic gradient-based algorithm (Adam \cite{kingmaAdamMethodStochastic2017}).
As the gradients of the variational lower bound $\mL_{\theta, \phi}$ with respect to the model parameters are intractable, we use the usual well-established Monte-Carlo methods for their estimation.

For the decoder parameters $\theta$, the gradient is estimated as
the sample gradient of the conditional log-likelihood with the latent $\bz$ and $\bc$ sampled from the approximate posterior
\begin{align}
\nabla_\theta \mL_{\theta, \phi}(\bx)
\approx \nabla_\theta \log p_\theta(\bx | \bz, \bc), 
\quad (\bz, \bc) \sim q_\phi(\bz, \bc | \bx)
\end{align}

For the encoder parameters $\phi$, we use the pathwise gradient estimators \cite{mohamedMonteCarloGradient2019} based on the standard location-scale $\bz = f_\phi(\tilde{\bz})$ and Gumbel-Softmax \cite{jangCategoricalReparameterizationGumbelSoftmax2017} $\bc = g_\phi(\tilde{\bc})$ reparametrizations with the auxiliary $\tilde{\bz} \sim \mN(0, 1)$ sampled from the standard normal and $\tilde{\bc}$ sampled from the Gumbel(0,1) distribution.
\begin{align}
\nabla_\phi \mL_{\theta, \phi}(\bx) 
\approx &
\nabla_\phi \log p_\theta(\bx, f_\phi(\tilde{\bz}), g_\phi(\tilde{\bc})) \nn
& - \nabla_\phi \log q_\phi(f_\phi(\tilde{\bz}), g_\phi(\tilde{\bc}) | \bx) \nn
& \tilde{\bz} \sim \mN(0, 1), \ \tilde{\bc} \sim \text{Gumbel}(0,1)
\end{align}

Finally, the gradients with respect to the parameters $\varphi$ of the conditional prior are estimated alongside the gradients of the decoder under the same sampling of the latents.
\begin{align}
\nabla_{\varphi} \mL_{\theta, \phi}(\bx)
\approx - \nabla_{\varphi} \log p_\varphi(\bz | \bc), \quad (\bz, \bc) \sim q_\phi(\bz, \bc | \bx) 
\end{align}

\subsubsection{Analysis of the objective}\label{sec:condVAE_objectiveAnalysis}

The $\KL$ divergence in term $B$ of the objective \eqref{eq:condVaeElbo} has important regularization effects on the model learning.
We expand on the discussion of these in the standard VAE objective \eqref{eq:vaeElbo} from \cite{zhaoInfoVAEInformationMaximizing2017} to analyse our more complex model formulation.

There are two major issues that optimizing the reconstruction term $A$ of the objective \eqref{eq:condVaeElbo} in isolation could cause.
First, the model could completely ignore the categorical component of the latent representation $\bc$ by encoding all the data points $\bx$ into a single category with a probability $q_\phi(c_k | \bx) = 1$.
All the variation in the data $\bx$ would then be captured within the continuous component of the latent representation through the single continuous posterior $q_\phi(\bz | \bx, c_k)$.
While this would not diminish the ability of the model to reconstruct the observed data and therefore would not decrease the reconstruction part of the objective $A$, it would degrade the generative properties of our model.
Specifically, with all the data clusters pushed into a single categorical component and distributed within the continuous latent space, we would have no leverage for generating samples from the individual data distributional components which is one of the major requirements for our method.
This pathological case is essentially equivalent to learning with the standard VAE.

Second, as discussed in \cite{zhaoInfoVAEInformationMaximizing2017}, maximizing the log-likelihood in $A$ naturally pushes the continuous posteriors to be concentrated around their means in disjoint parts of the continuous latent space with variances tending to zero.
For such posteriors, the model could learn very specific decoding yielding very good reconstructions with very high log-likelihoods $p_\theta(\bx | \bz, \bc)$.
However, the generations would again suffer as the prior used for the ancestral sampling would not cover the same areas of the latent space as used during the inference.

To analyse the reguralization effect of term $B$ in the objective \eqref{eq:condVaeElbo} on the learning, we decompose the expected $\KL$ divergence into three terms (and a constant):
\begin{align}\label{eq:condVAE_KLDecomp}
\mE_{p(\bx)} 
\KL\left(q_\phi(\bz, \bc | \bx) || p_\varphi(\bz, \bc)\right)
& = \mI_q\left( (\bz, \bc), \bx \right) \nn
& + \mE_{q_\phi(\bc)} \KL\left(
q_\phi(\bz| \bc) || p_\varphi(\bz | \bc)\right) \nn
& - 
\mH\left(q_\phi(\bc)\right) + \log K \enspace .
\end{align}
The first is the mutual information of the composite latent variable $(\bz, \bc)$ and the data $\bx$ under the posterior distribution $q$.
Minimization of the $\KL$ divergence in \eqref{eq:condVaeElbo} pushes the mutual information between the two to be low and therefore prevents the overfitting of the latent representation to the training data described in the second point above.

The third term is the negative entropy of the marginal categorical posterior whose empirical evaluation over the data sample $\mS = \{\bx_i\}_{i=1}^n$ is often referred to as the \emph{aggregated posterior} \cite{makhzaniAdversarialAutoencoders2016a,tomczakVAEVampPrior2017}.
\begin{equation}\label{eq:condVAE_aggc}
q_\phi(\bc) = \mE_{p(\bx)} q_\phi(\bc | \bx) \approx \frac{1}{n} \sum_i^n q_\phi(\bc | \bx_i)
\end{equation}
The regularizer maximizes the entropy of this distribution thus encouraging the model to use evenly all the categories of the discrete latent code counteracting the pathological case of the first point above.

Finally, the middle term pushes the marginalized conditional posteriors of the continuous latent variable $\bz$ to be close to the priors conditioned on the corresponding categories.
It helps to distribute the variations in the data into the continuous component of the latent space in agreement between the inferential posteriors and the learned generative priors.
It does so for each latent category $c_k$ separately, putting more or less weights on the alignment as per the importance of the latent categories established through the categorical marginal posterior $q_\phi(\bc)$.
It is this term in the objective of our VAE formulation which safeguards the generative properties of the model by matching the inferential posteriors and the learned generative priors used in the ancestral sampling procedure for new data examples.




\section{Related work}\label{sec:RelatedWork}

Since their introduction in 2014 \cite{kingmaAutoEncodingVariationalBayes2014,rezendeStochasticBackpropagationApproximate2014} variational autoencoders have become one of the major workhorses for large-scale density estimation and unsupervised representation learning.
Multitudes of variations on and enhancements of the original design have been proposed in the literature.
These can broadly be categorised into 4 large groups (with significant overlaps as many methods mix multiple ideas to achieve the best possible performance).

First, it has been argued that optimizing the variational bound \eqref{eq:vaeElbo} instead of the intractable likelihood $p_\theta(\bx)$ inhibits the VAEs to learn useful latent representations both for data reconstructions and for downstream tasks.
Methods using alternative objectives aim to encourage the learning towards representations that are better aligned with the data (measured by mutual information), e.g. InfoVAE \cite{zhaoInfoVAEInformationMaximizing2017}, \cite{alemiFixingBrokenELBO2018}, or which separate important factors of variations in the data (disentangling), e.g. \cite{higginsBetaVAELearningBasic2016,kimDisentanglingFactorising2019,dupontLearningDisentangledJoint2018}.
Though these methods report good results on occasions, there seem to be little evidence that breaking the variational bound brings systematical improvements \cite{locatelloChallengingCommonAssumptions2019a,daiDiagnosingEnhancingVAE2019}.

For our model, the analysis in section \ref{sec:condVAE_objectiveAnalysis} suggests that our objective (which is a proper lower bound on the likelihood) encourages the encoding of the major source of variation, that of the originating mixture component, through the categorical variable without any extra alterations.
At the same time, it should be noted that our goal is not the interpretability of the learned representations nor their reuse outside the VAE model.
Our focus is on generations reflecting the cluster structure of the original data space.

Second, the simplifying conditional independence assumptions for the data dimensions factored into the simple Gaussian decoder $p_\theta(\bx | \bz)$ have been challenged in the context of modelling data with strong internal dependencies.
More powerful decoders with autoregressive architectures have been proposed for modelling images, e.g. PixelVAE \cite{gulrajaniPixelVAELatentVariable2016},  or sequentially dependent data such as speach and sound, e.g. VRNN \cite{chungRecurrentLatentVariable2015a}.
In our model, we use a hierarchical decoder $p_\theta(\bx | \bz, \bc)$ corresponding to the cluster-like structure we assume for the data space.
However, in this work we stick to the simple independence assumption for the data dimensions. 
Augmenting our method with stronger decoder should in principal be possible and is open for future investigation.

Third, the insufficient flexibility of the variational posterior $q_\phi(\bz | \bx)$ to approximate the true posterior $p_\theta(\bz | \bx)$ has led to proposals for more expressive posterior classes.
For example, a rather successful approach is based on chaining invertible transformations of the latent variable \cite{rezendeVariationalInferenceNormalizing2015,kingmaImprovedVariationalInference2016}.
While increasing the flexibility of the approximate posterior improves the modelling objective through better reconstructions, without accompanied enhancements of the prior it does not guarantee better generations.

This has been recognised and addressed by the fourth group of improvements which focuses on the model prior and which our method pursues.
For example, the authors in \cite{xuSphericalLatentSpaces2018,davidsonHypersphericalVariationalAutoEncoders2018} replace the distributional class of the prior (together with the posterior) by von Mises-Fisher distributions with potentially better characteristics for high-dimensional data with hyperspherical latent space.

More related to ours are methods that suggest to learn the prior.
The VLVAE \cite{chenVariationalLossyAutoencoders2017} uses the autoregressive flows in the prior which are equivalent to the inverse autoregressive flows in the posterior \cite{kingmaImprovedVariationalInference2016}.
The increased richness of the encoding and prior distributions leads to higher quality generations, however, the prior cannot be used to generate from selected parts of the data space as our model can.

The VampPrior \cite{tomczakVAEVampPrior2017} proposes to construct the prior as a mixture of the variational posteriors over a learned set of pseudo-inputs.
These could be interpreted as learned cluster prototypes of the data.
However, the model does not learn the importance of the components in the mixture, and it does not align the prior and posteriors at an individual component level as our model does.
Instead, it pushes the posteriors to align with the overall prior mixture which diminishes the models ability to correctly generate from the individual components of multimodal data.

The continuous-discrete decompositions of the latent space similar to ours have been used for data clustering through generative model in \cite{jiangVariationalDeepEmbedding2017} and \cite{dilokthanakulDeepUnsupervisedClustering2017}.
The first combines the VAE with a Gaussian mixture model through two stage procedure mimicking the independence assumptions in their inference model.
The latter assumes (conditional) independences in the generative and inference models and extends to a full Bayesian formulation through the use of hyper-priors.
Their complex  model formulation exhibits some over-regularization issues that, as the authors acknowledge, are challenging to control.

Options for freeing the distributional class of the latent representations through Bayesian non-paremetrics have been explored for example in \cite{nalisnickStickBreakingVariationalAutoencoders2017,goyalNonparametricVariationalAutoencoders2017,liLearningLatentSuperstructures2018}.
The learned structures in the latent representations greatly increase the  generative capabilities including also the (hierarchical) clustering ability.
However, this comes at a cost of complex models that are tricky to train in a stable manner. In contrast, our model is elegantly simple and easy to train.

\section{Empirical evaluation}\label{sec:Experiments}
We validate our new conditional prior (CP-VAE) model through experiments over synthetic data and two real-image datasets (MNIST \cite{Lecun98gradientBasedlearning} and Omniglot \cite{Lake1332}).
We compare the results with those produced by standard VAE (VAE), VAE with Mixture of Gaussian prior (MG) and VAE with VampPrior (VP) \cite{tomczakVAEVampPrior2017}.

Not to obfuscate the analysis of the benefits of our method by various tweaks in the model architecture, we use the same structure of the encoder and decoder networks for all the methods in all our experiments.
We set the dimensions of the continuous latent variable to 40,
we use simple feed-forward networks with two hidden layers of 300 units each for both the encoder and the decoder, we initialise the weights according to Glorots's method \cite{glorot2010understanding}, and we utilize the gating mechanism of \cite{dauphin2017language} as the element-wise non-linearity.

We trained all the models using ADAM optimizer \cite{kingmaAdamMethodStochastic2017} with learning rate $5 \times 10^{-4}$, mini-batches of size $100$, and early stopping based on the stability of the objective over a validation-set.
To avoid pathological local minima and numerical issues during training, we use a warm-up \cite{bowman-etal-2016-generating} of 100 epochs until the full weighting of the $\KL$ regularization in the objective.

For generating new data examples, we use the ancestral sampling strategy with the latent variables sampled from the respective prior distributions of each method.
In the simple VAE this is from the standard normal Gaussian $\bz \sim p(\bz) = \mN(\bz | 0, I)$.
In the MG model it is from the set of learned Gaussian components $\bz \sim p_\varphi(\bz) = \frac{1}{K} \sum_i^K \mN(z | \mu_k, \diag(\sigma_k))$ with equal weighting.
For VP it is from the mixture of variational posteriors $\bz \sim p_\varphi(\bz) = \frac{1}{K} \sum_i^K q_\phi(\bz | \bu_k)$ over the learned set of pseudo-inputs $\mU = \{ \bu_k\}_{i=1}^K$ which first have to be passed through the encoder network.
Four our method, we follow the two level-generative process described in section \ref{sec:condVAE_genProcess} where we use the learned conditional priors for each of the categories for sampling the continuous latent component $\bz \sim p_\varphi(\bz | \bc = c_k) = \mN(z | \mu_k, \diag(\sigma_k))$ and the empirical aggregated posterior \eqref{eq:condVAE_aggc} for the discrete component $\bc \sim q_\phi(\bc)$.

The implementation of our method together with the settings for replication of our experiments is available from our Bitbucket repository \url{https://bitbucket.org/dmmlgeneva/cp-vae/src/master/}.

\subsection{Synthetic data experiments}
In this section we demonstrate the effectiveness of the CP-VAE method through experiments over synthetic data.
We use a toy dataset with 50000 examples $\bx \in \mR$ generated from a Gaussian mixture with two equally weighted components $\bx \sim p(\bx) = \frac{1}{2} (N(0.1, 0.1) + N(1, 0.1))$.
This simple set-up allows us to better understand the strengths and weaknesses of the method in terms of its density estimation performance for a known and rather simple ground-truth data distribution.

Because in real-life problems the number of distributional clusters in the data (the number of mixture components) may not be known or even easy to estimate, we use two experimental set-ups:
\begin{itemize}
\item \textbf{known} number of components: discrete latent variable $\bc$ with 2 categories (corresponding to the ground-truth two mixture components)
\item \textbf{unknown} number of components: discrete latent variable $\bc$  with 150 categories
\end{itemize}

\begin{figure}[hbt!]
\centering
\subfigure[Real]{
\includegraphics[width=.130\textwidth]{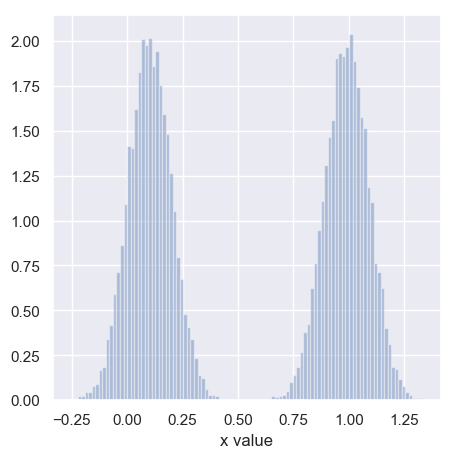}
}
\subfigure[CP 2]{
\includegraphics[width=.130\textwidth]{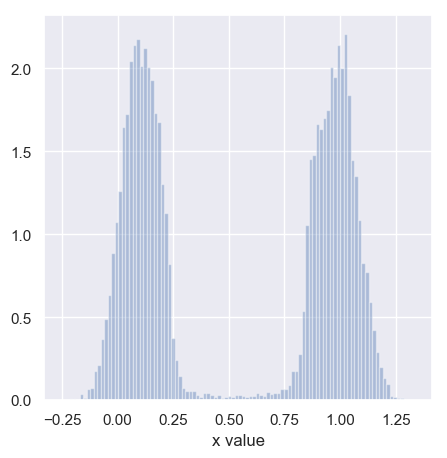}
}
\subfigure[CP 150]{
\includegraphics[width=.130\textwidth]{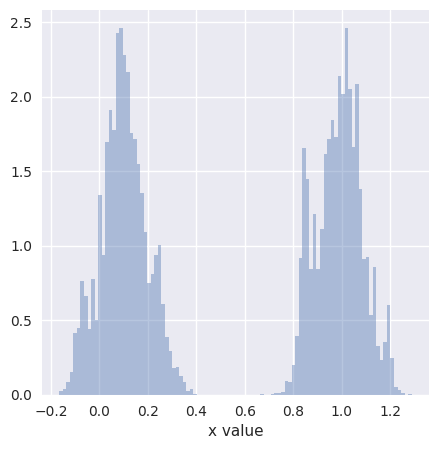}
}
\subfigure[VAE]{
\includegraphics[width=.130\textwidth]{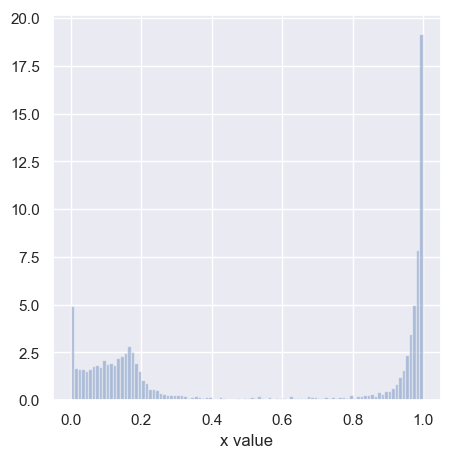}
}
\subfigure[MG 2]{
\includegraphics[width=.130\textwidth]{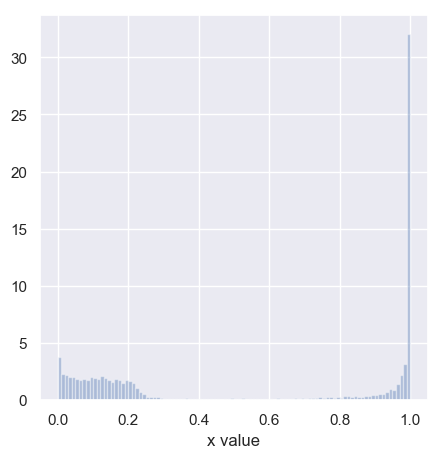}
}
\subfigure[MG 150]{
\includegraphics[width=.130\textwidth]{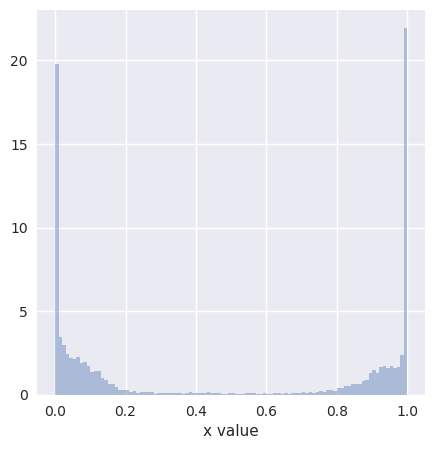}
}
\caption{Histograms of the data generated from the ground-truth and the learned distributions. Our CP method can recover the bi-modal structure of the data correctly irrespective of the number of categories used for the latent categorical component. Methods with priors not adapted to the learned disjoint representation fail.}
\label{fig_gaussian2_generations}
\end{figure}

In Figure \ref{fig_gaussian2_generations} we present histograms of data generated from the ground truth and the learned distributions.
As we can see, our method (CP) correctly recovers the bi-modal structure of the data for both of the set-ups.
This is important for practical utility of the method in situations where the domain knowledge does not provide an indication on the number of underlying generative clusters. With high enough number of categories within the discrete latent, our method can recover the correct multi-modal structure of the data.

As expected, methods which do not adjust their priors to the disjoint learned representation, such as the simple VAE or MG methods, cannot recover the multimodal structure of the data at generation time.

We further explore how our model handles the excess capacity within the categorical latent variable.
For this we focus on the 150-category case and we generate data by sampling the discrete latent variable a) from the marginal posterior $\bc \sim q_\phi(\bc)$, b) from the uniform prior $\bc \sim p(\bc) = \frac{1}{K}$.

\begin{figure}[hbt!]
\centering
\subfigure[$\bc \sim q_\phi(\bc)$]{
\includegraphics[width=.15\textwidth]{figures/gaussian_data/mine_generations_test_c150}
}
\subfigure[$\bc \sim p(\bc) = \frac{1}{K}$]{
\includegraphics[width=.15\textwidth]{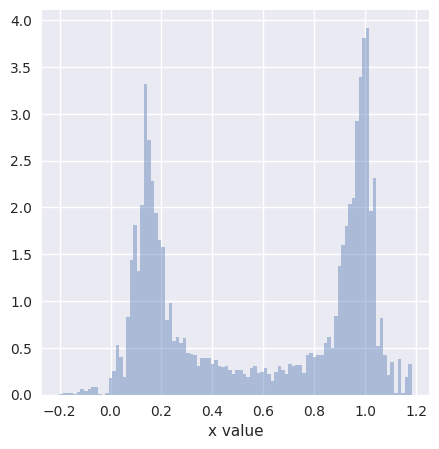}
}
\caption{The CP-VAE learns to ignore the excess capacity of the disjoint latent space by assigning near-zero probability to some of the categories in the discrete latent space.
These parts of the latent space are ignored for the reconstructions and by sampling the categorical variable from the marginal posterior $q_{\phi}(c)$ can correctly be ignored also for the generations.}
\label{fig_gaussian2_marginal}
\end{figure}

Comparing the two in Figure \ref{fig_gaussian2_marginal}, we see that unlike the generations sampled from the marginal posterior, the generations from the uniform prior display some mixing artifacts.
This suggests that our model learns to ignore the excess capacity by assigning low marginal probabilities $q_\phi(c_k) \approx 0$ to some of the categories.
The continuous latent representations corresponding to these parts of the disjoint latent space are irrelevant for both the reconstructions and the generations due to our weighted $\KL$ formulation in $B2$ of equation \eqref{eq:condVaeKLs}.

\subsection{Real data experiments}
For the real-data experiments we use to image datasets, 
MNIST \cite{Lecun98gradientBasedlearning} and Omniglot \cite{Lake1332}, commonly used for evaluation of generative models.
We use the dynamically binarized versions of the datasets with the following train-validation-test splits: for MNIST 50000-10000-10000, for Omniglot 23000-1345-8070.

We first demonstrate the benefits of learning the prior distribution $p_\varphi$  on the generative performance of our CP-VAE method.
For this we fix the number of categories within the discrete latent variable to 1. 
This is equivalent to learning the simple VAE with only the continuous latent representation but a learnable prior and the corresponding objective 
\begin{equation}
\mL_{\theta, \phi}(\bx) = \mE_{q_\phi(\bz  | \bx)} \log p_\theta(\bx | \bz) -  \KL\left( q_\phi(\bz| \bx) || p_\varphi(\bz)\right)
\end{equation}

\begin{figure}[hbt!]
\centering
\subfigure[Real]{
\includegraphics[width=.130\textwidth]{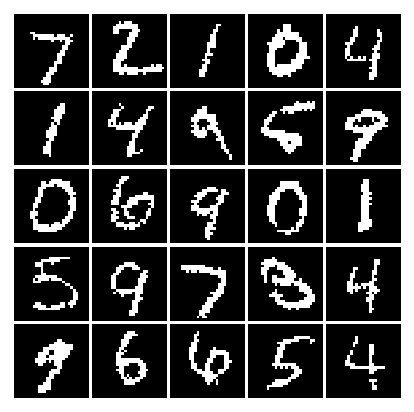}
}
\subfigure[CP 1]{
\includegraphics[width=.130\textwidth]{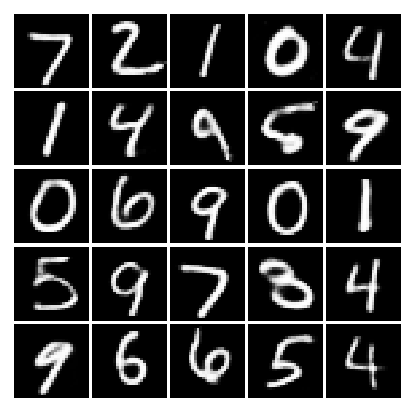}
}
\subfigure[VAE]{
\includegraphics[width=.130\textwidth]{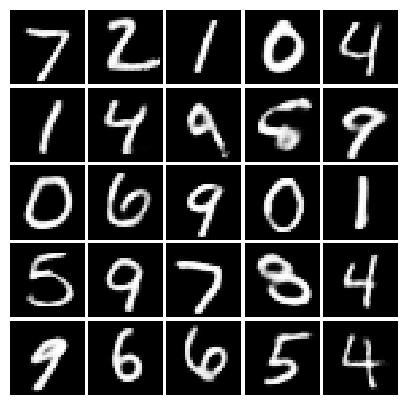}
}
\caption{Data reconstructions from models without a categorical structure in the latent space are similarly good with $p_\varphi(\bz)$ and without $p(\bz)$ using the learnable prior.}
\label{fig_MNISTc1_rec}
\end{figure}

\begin{figure}[hbt!]
\centering
\subfigure[CP1]{
\includegraphics[width=.130\textwidth]{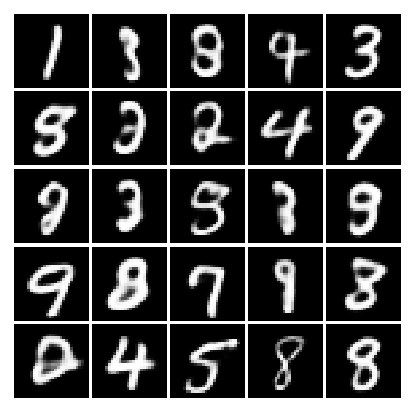}
}
\subfigure[VAE]{
\includegraphics[width=.130\textwidth]{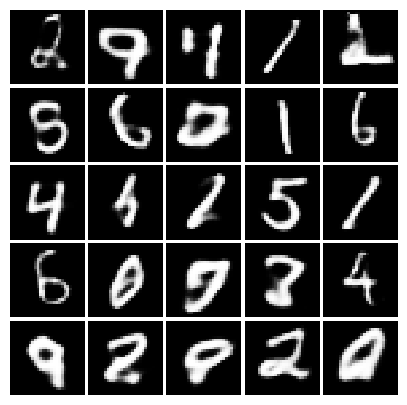}
}
\caption{New data generations from models without a categorical structure in the latent space are of clearly higher quality for the CP1 adapting the prior $p_\varphi(\bz)$ to the learned latent space than for the standard VAE with fixed prior $p(\bz)$.}
\label{fig_MNISTc1_gen}
\end{figure}

In Figures \ref{fig_MNISTc1_rec} and \ref{fig_MNISTc1_gen} we compare the reconstructions produced by our CP-VAE model with only a single category in the latent space to the standard VAE without a learnable prior.
The visual quality of the data reconstructions for both of the models is very high.
However, the quality of generations (Figure \ref{fig_MNISTc1_gen}) is clearly much better when the model adapts the prior to the learned latent space (CP1).

\begin{figure}[hbt!]
\centering
\subfigure{
\includegraphics[width=.08\textwidth]{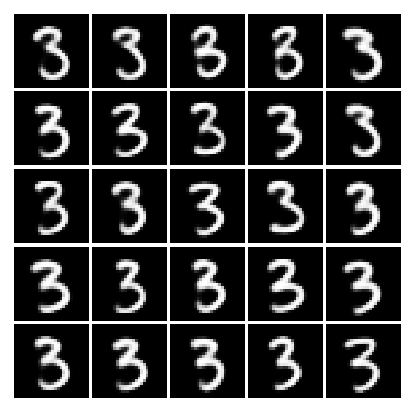}
}
\subfigure{
\includegraphics[width=.08\textwidth]{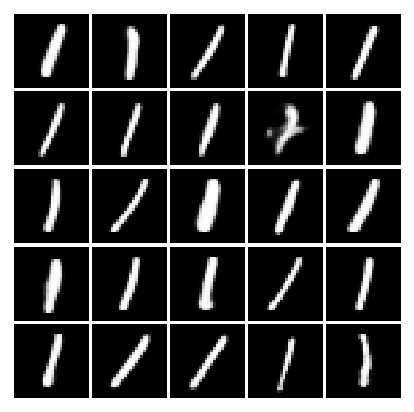}
}
\subfigure{
\includegraphics[width=.08\textwidth]{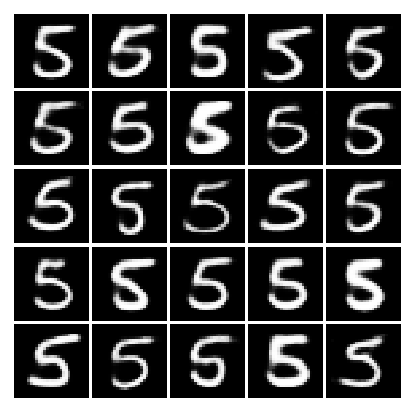}
}
\subfigure{
\includegraphics[width=.08\textwidth]{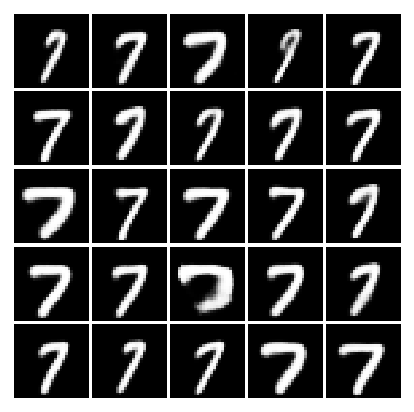}
}
\subfigure{
\includegraphics[width=.08\textwidth]{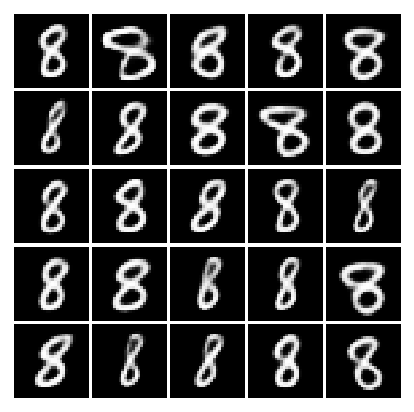}
}
\subfigure{
\includegraphics[width=.08\textwidth]{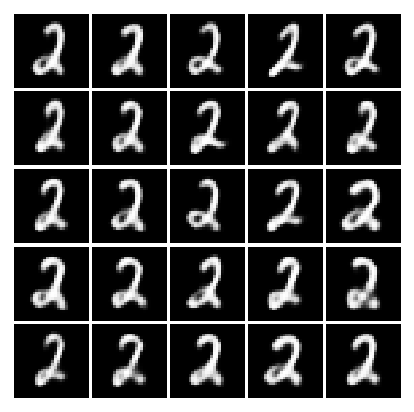}
}
\subfigure{
\includegraphics[width=.08\textwidth]{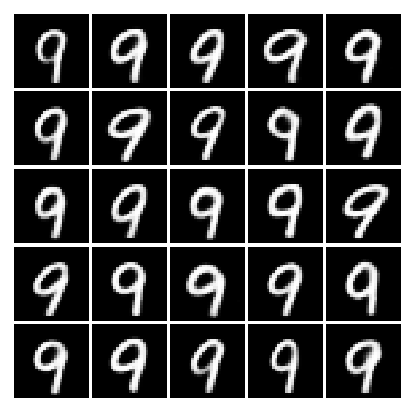}
}
\subfigure{
\includegraphics[width=.08\textwidth]{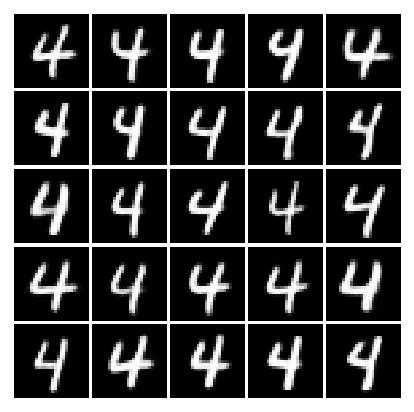}
}
\subfigure{
\includegraphics[width=.08\textwidth]{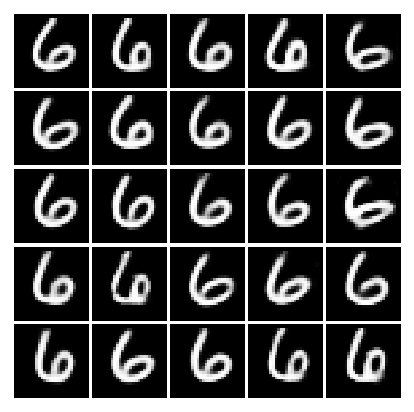}
}
\subfigure{
\includegraphics[width=.08\textwidth]{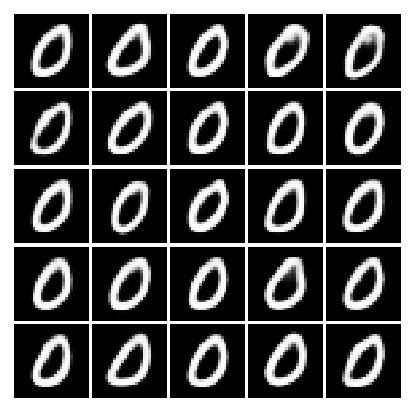}
}
\caption{New variations of individual MNIST digits generated by our CP-VAE model with a latent discrete code with 10 categories. Digits in the same subplot were generated from the same discrete category.}
\label{fig_MNIST_conditional_generations}
\end{figure}

We next examine the ability of our model to generate new examples from the underlying distributional clusters at will.
For this we trained our model over the MNIST data with 10 categories in the discrete latent variable.
In Figure \ref{fig_MNIST_conditional_generations} we show that our model learned to use the categorical variable for encoding the changes in the digits.
The examples in each of the subplots were generated by fixing the discrete latent variable to one category and sampling the continuous latent from the corresponding learned prior.
We see that the learned discrete encoding accurately captures the main categories of the data. 
In fact, rather remarkably, though are model was trained in a purely unsupervised manner, if we use the learned discrete code as a classification output we achieve $94.6\%$ classification accuracy.


Finally, we compare the performance of our CP-VAE to a number of baselines varying also the size of the categorical variable.
We experiment with $\{10, 150,  500 \}$ categories for MNIST and $\{50,  500 \}$ categories for Omniglot.
The methods we compare to are the simple VAE (VAE) and the following methods from  \cite{tomczakVAEVampPrior2017}: VAE with Mixture of Gaussians prior (MG), VAE with VampPrior (VP), hierarchical two-layerd VAE with VampPrior (HVP), and hierarchical two-layerd VAE with simple fixed prior (HVAE).
For the VP and MG methods we use the same numbers of pseudo-inputs and mixture components as the number of the latent categories and for the two layers models we use 40 latent variables at each layer.

\begin{table}
\ra{1.3}
\begin{tabular}{@{}rrrrcrr@{}}\toprule
& \multicolumn{3}{c}{MNIST} & \phantom{abc}& \multicolumn{2}{c}{$Omniglot$} \\ \cmidrule{2-4} \cmidrule{6-7} 
& $c = 10 $ & $c=150$ & $c=500$ && $c=50$ & $c=500$  \\ \midrule
\\
CP & \textbf{84.76} & 86.84&  88.63&& 114.31 & 114.14  \\
VAE & 88.75 &---&  --- && 115.06 &--- \\
MG & 89.43 & 88.96& 88.85&& 116.31& 116.12  \\
VP & 87.94 & 86.55 & 86.07 && 114.01 & 113.74 \\ 
HVAE & 86.7&---&--- && 110.81 &--- \\
HVP & 85.90 &\textbf{85.09}  &\textbf{85.01}  && \textbf{110.50} &  \textbf{110.16}\\ \bottomrule
\end{tabular}
\caption{Comparison of negative variational lower bounds for the different methods over the test data sets.}
\label{table:elbo}
\end{table}

We summarize the numerical results in terms of the negative variational lower bound calculated over the test data in Table \ref{table:elbo}.
Our model achieves lower bounds when compared to VAE and MG in all the cases.
It also performs the best of all the tested methods for the MNIST dataset with 10 latent categories.
Otherwise, the VampPrior and especially the hierarchical VampPrior method seems to perform consistently the best.
However, as explained in section \ref{sec:RelatedWork}, good values of the variational lower bound objective do not guarantee good generations and good control over the distributional clusters.

\begin{figure}[hbt!]
\centering
\subfigure[CP10]{
\includegraphics[width=.08\textwidth]{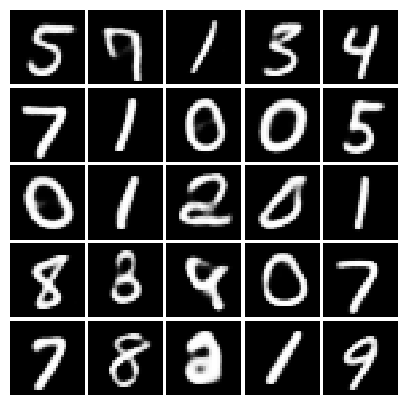}
}
\subfigure[VP10]{
\includegraphics[width=.08\textwidth]{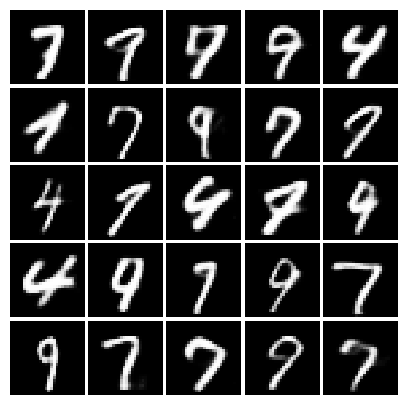}
}
\subfigure[HVP10]{
\includegraphics[width=.08\textwidth]{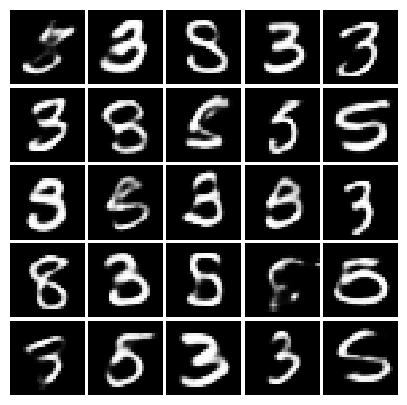}
}
\subfigure[MG10]{
\includegraphics[width=.08\textwidth]{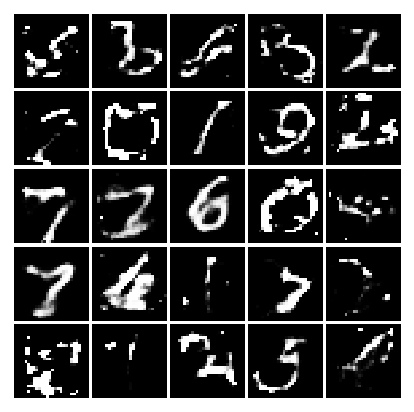}
}
\subfigure[VAE]{
\includegraphics[width=.08\textwidth]{figures/mnist/vae_generations}
}

\subfigure[CP150]{
\includegraphics[width=.08\textwidth]{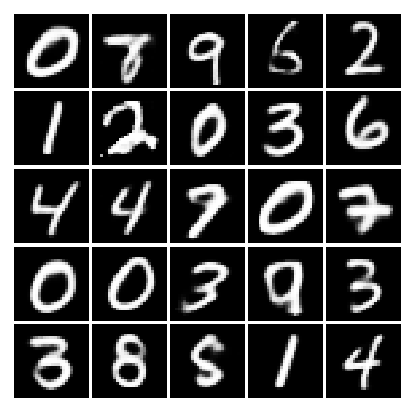}
}
\subfigure[VP150]{
\includegraphics[width=.08\textwidth]{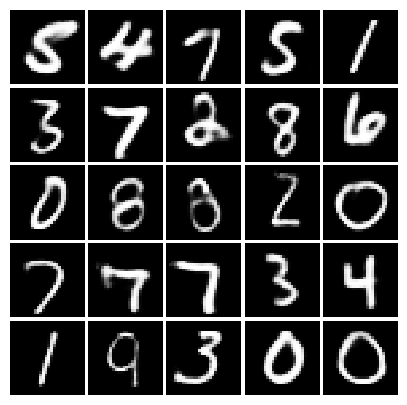}
}
\subfigure[HVP150]{
\includegraphics[width=.08\textwidth]{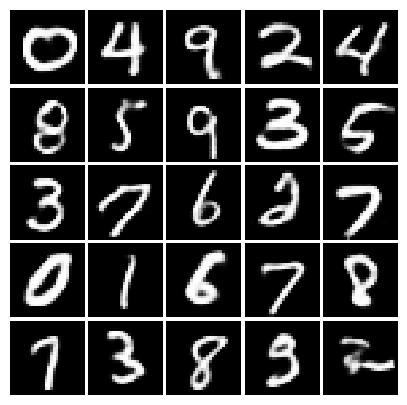}
}
\subfigure[MG150]{
\includegraphics[width=.08\textwidth]{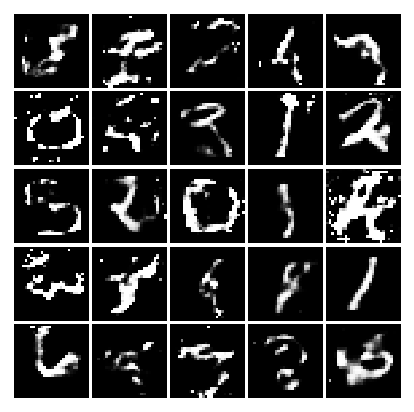}
}
\subfigure[HVAE]{
\includegraphics[width=.08\textwidth]{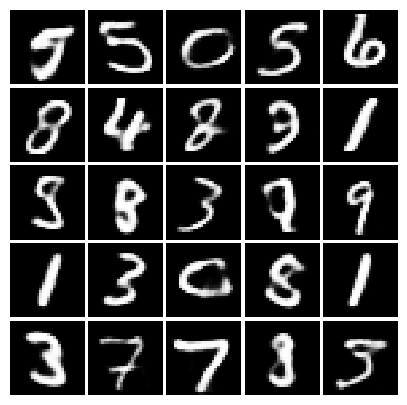}
}

\subfigure[CP500]{
\includegraphics[width=.08\textwidth]{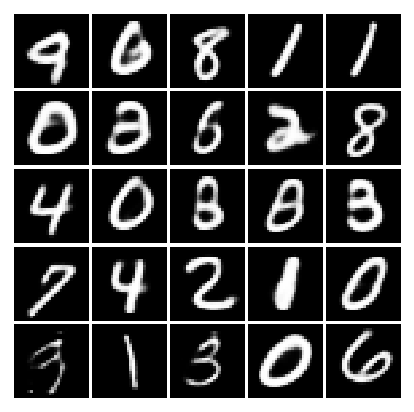}
}
\subfigure[V500]{
\includegraphics[width=.08\textwidth]{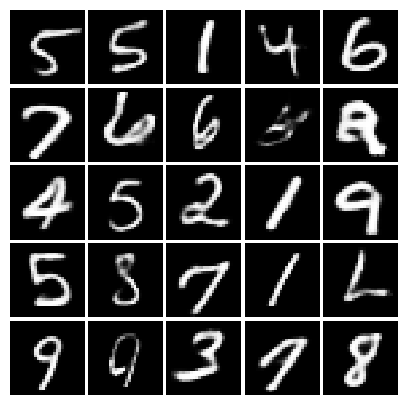}
}
\subfigure[HV500]{
\includegraphics[width=.08\textwidth]{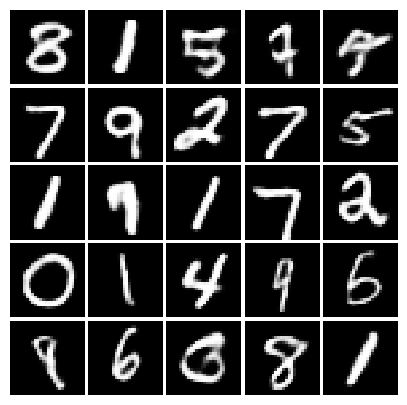}
}
\subfigure[MG500]{
\includegraphics[width=.08\textwidth]{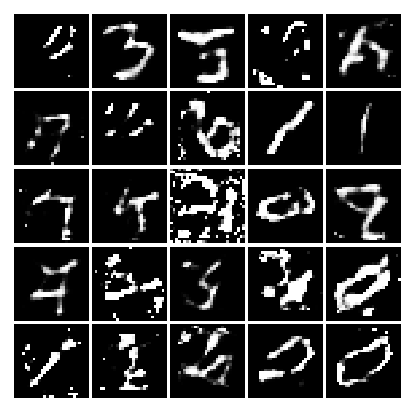}
}

\caption{New data examples from the MNIST dataset generated by the various methods with increasing number of the prior components.}
\label{fig_MNIST}
\end{figure}

\begin{figure}[hbt!]
\centering
\subfigure[CP50]{
\includegraphics[width=.08\textwidth]{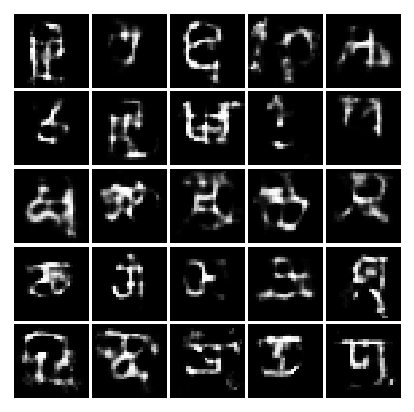}
}
\subfigure[VP50]{
\includegraphics[width=.08\textwidth]{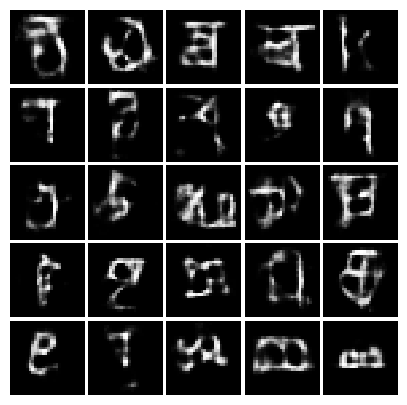}
}
\subfigure[HVP50]{
\includegraphics[width=.08\textwidth]{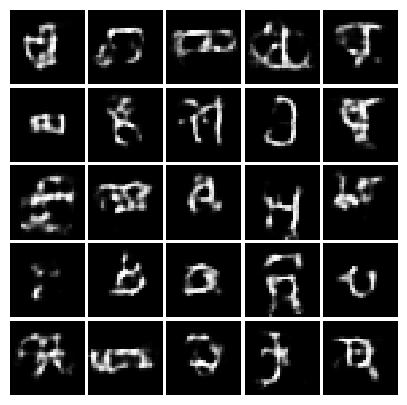}
}
\subfigure[MG50]{
\includegraphics[width=.08\textwidth]{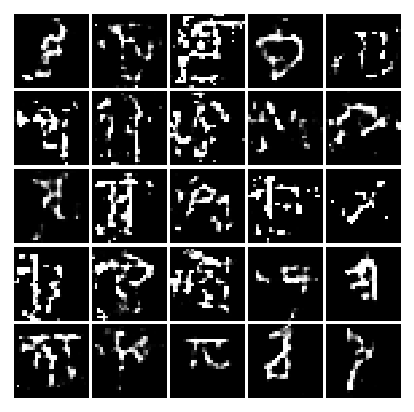}
}
\subfigure[VAE]{
\includegraphics[width=.08\textwidth]{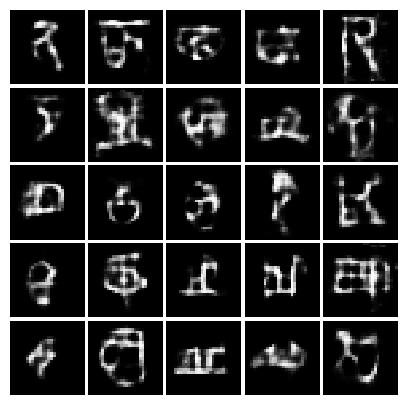}
}

\subfigure[CP500]{
\includegraphics[width=.08\textwidth]{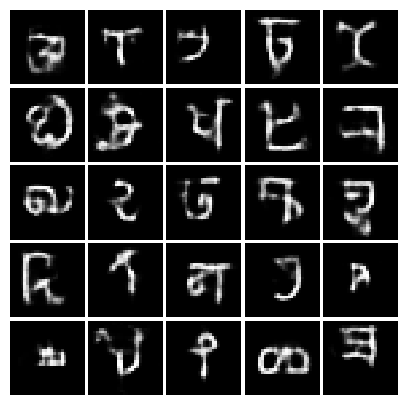}
}
\subfigure[VP500]{
\includegraphics[width=.08\textwidth]{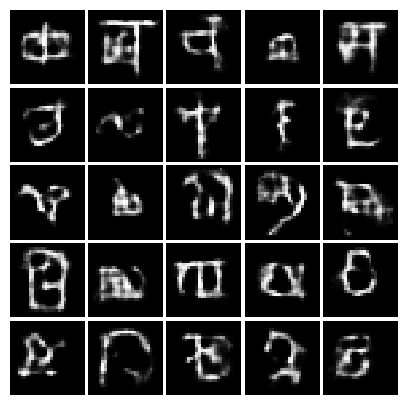}
}
\subfigure[HVP500]{
\includegraphics[width=.08\textwidth]{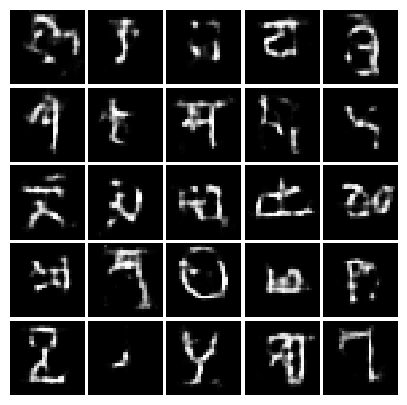}
}
\subfigure[MG500]{
\includegraphics[width=.08\textwidth]{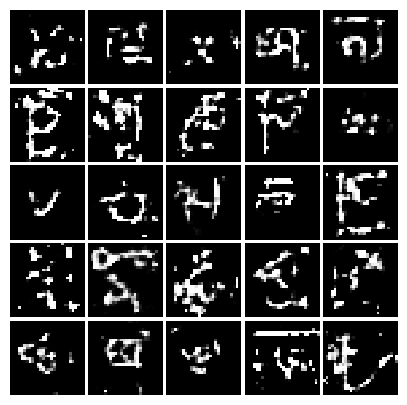}
}
\subfigure[HVAE]{
\includegraphics[width=.08\textwidth]{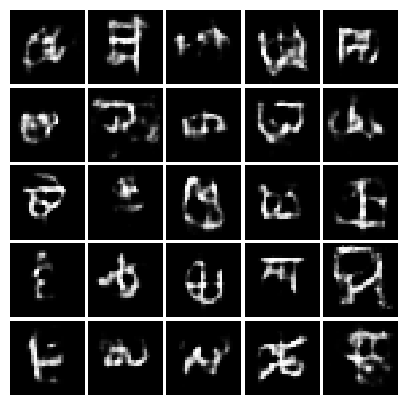}
}
\caption{New data examples from the Omniglot dataset generated by the various methods with increasing number of the prior components.}
\label{fig_omniglot}
\end{figure}

We present the new data examples generated by the various methods in Figure \ref{fig_MNIST} and \ref{fig_omniglot} for the MNIST and Omniglot data respectively.
Our model is able to generate consistently good quality new samples for both the datasets irrespective of the number of latent categorical components.
In contrast, all of the other methods (all VampPrior variations including the two-layer hierarchical, and the MG) fail to generate quality examples with only 10 components within the prior.
They also seem to collapse to generating examples only from a few digits which suggest an important lack of flexibility available for the generations.
As the number of components (pseudo-inputs) in the prior mixture increases, the VampPrior generations tend to improve, with the hierarchical version of the method systematically outperforming the simple VP version.

\newpage
\section*{APPENDIX}\label{sec:Appendix}

Proof of equation \eqref{eq:condVaeKLs}

\begin{multline*}
\underbrace{\KL\left(q_\phi(\bz, \bc | \bx) || p_\varphi(\bz, \bc)\right)}_B = \\
= \mE_{q_\phi(\bz|\bx, \bc) q_\phi(\bc|\bx)}
\log
\frac{q_\phi(\bc|\bx)}{p(\bc)}
\frac{q_\phi(\bz|\bx, \bc)}{p_\varphi(\bz | \bc)}
= \\
=
\underbrace{\KL\left(q_\phi(\bc| \bx) || p(\bc)\right)}_{B1}
+ \underbrace{\mE_{q_\phi(\bc|\bx)} \KL\left( q_\phi(\bz| \bx, \bc) || p_\varphi(\bz | \bc)\right)}_{B2}
\end{multline*}
Proof of equation \eqref{eq:condVae_KLB1}
\begin{align*}
\underbrace{\KL\left(q_\phi(\bc| \bx) || p(\bc)\right)}_{B1} & =
\mE_{q_\phi(\bc| \bx)} \left[ \log q_\phi(\bc| \bx) - \log K^{-1}\right] \nn
& =
- \mH\left(q_\phi(\bc|\bx)\right) + \log K
\end{align*}
Proof of equation \eqref{eq:condVAE_KLDecomp}
\begin{multline*}
\mE_{p(\bx)} 
\KL\left(q_\phi(\bz, \bc | \bx) || p_\varphi(\bz, \bc)\right) = \nn
\mE_{p(\bx)} \KL\left(q_\phi(\bc| \bx) || p(\bc)\right) \nn
+ \mE_{p(\bx)} \mE_{q_\phi(\bc|\bx)} \KL\left( q_\phi(\bz| \bx, \bc) || p_\varphi(\bz | \bc)\right)
= \nn
= 
\mE_{q_\phi(\bx, \bc)} 
\left[ 
\log \frac{q_\phi(\bc| \bx)}{q_\phi(\bc)}
+ \log \frac{q_\phi(\bc)}{p(\bc)}
\right] \nn
+
\mE_{q_\phi(\bx, \bz, \bc)} 
\left[ 
\log \frac{q_\phi(\bz, \bc| \bx)}{q_\phi(\bc | \bx) q_\phi(\bz | \bc)}
+ \log \frac{q_\phi(\bz| \bc)}{p_\varphi(\bz | \bc)}
\right] \nn
=
\KL\left(q_\phi(\bc) || p(\bc) \right)
+ \mE_{q_\phi(\bc)} \KL\left(
q_\phi(\bz| \bc) || p_\varphi(\bz | \bc)\right) \nn
+ \mE_{q_\phi(\bx, \bz, \bc)} 
\log \frac{q_\phi(\bz, \bc| \bx)}{q_\phi(\bz| \bc)q_\phi(\bc)} \nn
=
\log K - \mH(q_\phi(\bc))
+ \mE_{q_\phi(\bc)} \KL\left(
q_\phi(\bz| \bc) || p_\varphi(\bz | \bc)\right) \nn
+ \mE_{q_\phi(\bx, \bz, \bc)} 
\log \frac{q_\phi(\bz, \bc| \bx)}{q_\phi(\bz, \bc)} \nn
=
\log K - \mH(q_\phi(\bc)) \nn
+ \mE_{q_\phi(\bc)} \KL\left(
q_\phi(\bz| \bc) || p_\varphi(\bz | \bc)\right) +
\mI_q\left( (\bz, \bc), \bx \right) \nn
\end{multline*}

\bibliography{ecai2020_condPrior_v10}

\end{document}